# DeepXPalm: Tilt and Position Rendering using Palm-worn Haptic Display and CNN-based Tactile Pattern Recognition

Miguel Altamirano Cabrera[1], Oleg Sautenkov[1], Jonathan Tirado[1], Aleksey Fedoseev[1], Pavel Kopanev[1], Hiroyuki Kajimoto[2], Dzmitry Tsetserukou[1]

*Abstract*— Telemanipulation of deformable objects requires high precision and dexterity from the users, which can be increased by kinesthetic and tactile feedback. However, the object shape can change dynamically, causing ambiguous perception of its alignment and hence errors in the robot positioning. Therefore, the tilt angle and position classification problem has to be solved to present a clear tactile pattern to the user. This work presents a telemanipulation system for plastic pipettes consisting of a multi-contact haptic device LinkGlide to deliver haptic feedback at the users' palm and two tactile sensors array embedded in the 2-finger Robotiq gripper. We propose a novel approach based on Convolutional Neural Networks (CNN) to detect the tilt and position while grasping deformable objects. The CNN generates a mask based on recognized tilt and position data to render further multi-contact tactile stimuli provided to the user during the telemanipulation. The study has shown that using the CNN algorithm and the preset mask, tilt, and position recognition by users is increased from 9.67% using the direct data to 82.5%.

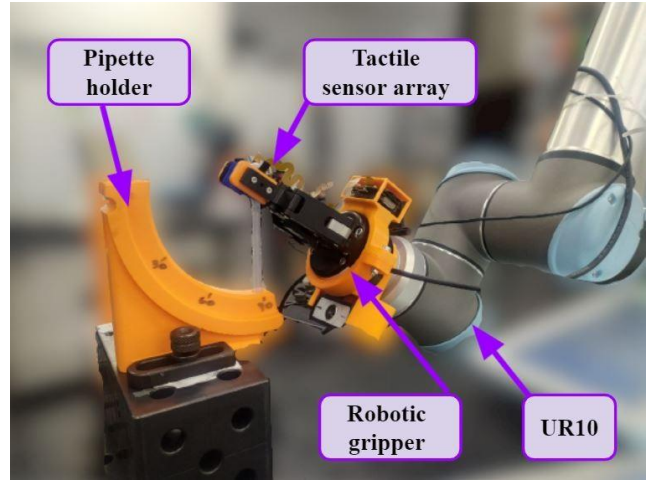

Fig. 1. Tilt angle recognition by the DeepXPalm system with high density tactile sensor array.

## I. INTRODUCTION

The increasing number of teleoperation and telexistence systems in recent years has opened a demand for reliable feedback on the position and orientation of manipulated objects to achieve dexterous interaction with them.

Many studies have started implementing tactile displays to provide high-fidelity feedback. Tachi et al. [1] introduced the concept of a highly immersive and mobile telexistence system TELESAR, which includes an autostereoscopic 3D display for visual feedback, wearable tactile and thermal display. Several wearable haptic devices were suggested to achieve a highly immersive VR experience, with a significant focus on the human fingertips due to the high density of Rapidly Adapting (RA) tactile receptors in this area. For example, in the works of Mengoni et al. [2], and Peruzzini et al. [3] electro-stimulation was applied to deliver the sensation of roughness, slickness, and texture coarseness of materials (e.g., wood, paper, fabric) to the fingertip. Pacchierotti et al. [4] developed the teleoperation system that performs both kinesthetic and vibrotactile feedback at the pen-shaped handle to decrease the targeting error of the robot and the orientation error of its tool. A wearable 3-DoF fingertip haptic displays was suggested by Gabardi et al. [5] for shape rendering in virtual reality. The multimodal stimulation was further explored by Yem et al. [6] with FinGAR, a wearable tactile device using mechanical and electrical stimulation for fingertip interaction with the virtual world. Tirado et al. [7] proposed a tactile sharing system ElectroAR for remote training of human hand skill, with a tactile sensing glove on the follower's side that records the pressure data about grasped objects and an electro-tactile stimulation glove on the remote side.

However, while the density of the receptors in the human fingertips of 141 *units/cm*$^2$ is higher than their density in the palm of 25 *units/cm*$^2$, the overall number of receptors on the palm is compensated by its larger area, having a total of 30% of all the RA receptors of the glabrous skin of the hand [8]. The investigation on the user's palm sensitivity, performed by Altamirano et al. [9], revealed distinguishable active areas at the palm during the interaction with surfaces at different forces and a high recognition rate of tactile patterns representing these applied forces.

While the palm area can potentially be utilized for the high-fidelity haptic feedback, few haptic devices were aimed at this area of the human hand. Martínez et al. [10] developed a vibrotactile glove for virtual reality with twelve vibrotactile actuators on the fingertips and palm of the user. Altamirano et al. [11] developed a multi-contact wearable haptic display with inverted five-bar linkages. The linkage structure was also utilized by Son et al. [12] in combination with an exoskeleton glove for the tactile perception of large objects.

[1]The authors are with the Space Center, Skolkovo Institute of Science and Technology (Skoltech), 121205 Bolshoy Boulevard 30, bld. 1, Moscow, Russia. {miguel.altamirano, oleg.sautenkov jonathan.tirado, aleksey.fedoseev, pavel.kopanev, d.tsetserukou}@skoltech.ru

[2]The author is with The University of Electro-Communications, Tokyo, Japan. kajimoto@kaji-lab.jp

Fedoseev et al. [13] proposed a linkage array for tactile rendering with an encountered-type haptic display. Fani et al. [14] implemented the system where mechanotactile feedback was used at several point on human arm to improve the telemanipulation of a robotic arm in a dexterous operation.

In this paper, we present a novel CNN-based telemanipulation system DeepXPalm for collaborative robots that consists of a wearable haptic device and two tactile sensors array embedded in a 2-finger Robotiq gripper. We propose an approach based on CNN to detect deformable objects' tilt and position.

The use of pipettes is imminent in laboratories due to the ongoing pandemic. This work would like to contribute to the challenge of the fulfillment of COVID tests in remote places employing robotic systems increasing the medical staff's safety. In this study, a Pasteur Pipette for liquid transfer (total length: 150 $mm$, total capacity: 7 $ml$, material: low-density polyethylene) was chosen because of its large deformation.

Two experiments were carried out to assess the performance of the approach. First, we evaluated the human perception when downsize data is rendered by the haptic device to the users' palm. Second, we evaluates the tilt and position recognition using the mask generated by the CNN algorithm.

## II. DeepXPalm System Overview

### A. System Architecture

The exterior of the system and the overall architecture are depicted in Fig. 1 and Fig. 2, respectively.

A 2-finger gripper from Robotiq mounted on the end-effector of the UR3e robotic arm was used to grasp the plastic pipettes. The tactile sensing device is equipped with two sensor grids, one on each fingertip of the gripper, thus allowing measurement of the local pressure at each point on the gripper's fingertip.

The wearable palm-worn tactile display LinkGlide [11] is used to deliver multi-contact stimuli at the user's palm. The array of inverted five-bar linkages generates three independent contact points to cover the whole palm area.

The measured pressure in the gripper is then rendered accordingly on the user's palm using multi-contact tactile stimulation by LinkGlide.

All processes were supervised by a PC (Intel i7-7700HQ CPU @2.8GHz x 8, 15 GB of RAM, GeForce GTX 1070, running Ubuntu 18.04.5), which also executes the convolutional neural network (CNN) process and maintains communication between the output and input systems. The sample time of the whole process is set at 16.67 $ms$ (60 $Hz$). During this time, the system senses the real objects, analyzes the sensed data, and delivers the tactile stimuli to the user's palm.

The force sensor reads the values from the sensor grid over the USART interface and passes the measurements to the CNN and tactile display. The CNN recognizes the position and the orientation of the object and conveys the corresponding mask to the tactile display. Subsequently, the tactile display uses both actual measurements from the sensor and the orientation mask. The mask is used to suppress random noise coming from irrelevant pins, thus enhancing the overall perception and pattern recognition capabilities of the user.

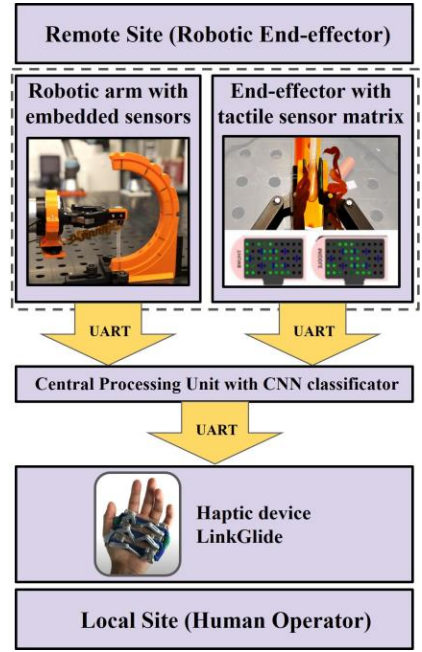

Fig. 2. The overall architecture of the telemanipulation system. Green arrows define a hardware integration; blue arrows define a control signal; yellow arrows define a feedback loop.

### B. Palm Recognition Rate

The system transmits to a user the position and the orientation of the object. We estimated twelve patterns for tactile rendering, four different angles (0, 45, 90, and 135 degrees) in three different positions each (center, left, and right for 45, 90, and 135 degrees, and center, up, and down for 0 degrees). The study developed by Altamirano et al. [9] showed the human hand recognition rate of tactile patterns. The experiment on tactile pattern detection revealed a high recognition rate 84.29% of a human palm. Based on this research, we decided to deliver the patterns onto a human's palm.

### C. High Fidelity Tactile Sensory Subsystem

The 2-finger robotic gripper is embedded with high-density tactile sensor arrays [15]. One sensor array is attached to each fingertip. A single electrode array can sense a frame area of 5.8 $cm^2$ with a resolution of 100 points per frame. The sensing frequency equals 120 $Hz$. The range of force detection per point is of 1–9 $N$. Thus, the robot detects the pressure applied to solid or flexible objects grasped by the robotic fingers with a resolution of 200 points (100 points per finger).

The data collected from the sensor arrays are processed by a downsize algorithm or by a CNN-based approach to evaluate non-deterministic data (see Section III). The neural network's objective is to estimate the tilt orientation of grasped objects.

## D. LinkGlide Haptic Device

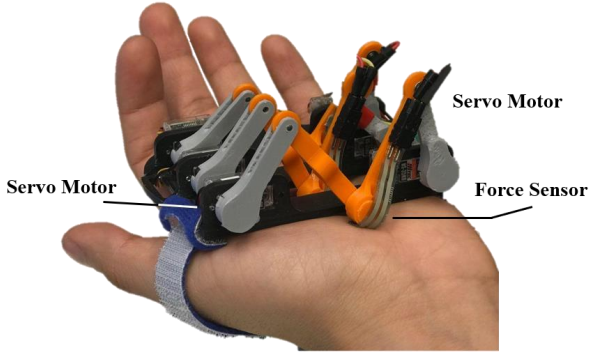

Fig. 3. LinkGlide haptic device.

The LinkGlide device provides the sense of touch at three different points in the palm of the user where multimodal stimuli can be delivered. The proposed device is based on LinkTouch technology [16]. Three 2-DoF inverted five-bar mechanisms, distributed in parallel planes, deliver object detection and manipulation sensation, produce the sliding force, and multi-contact state at the palm. Each of the mechanisms has two servo motors PowerHD DSM44, which angles provide the planar position of the single contact point between palm and linkage. Therefore, three contact points can be created, i.e., $C_1$, $C_2$, $C_3$. The linkage configuration of LinkGlide is presented in Fig. 4.

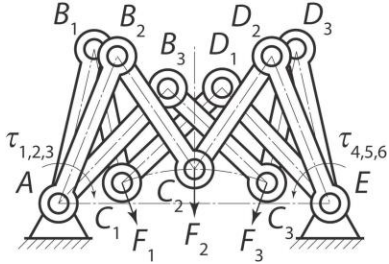

Fig. 4. Linkage configuration of LinkGlide.

The equation of $\tau_{an}$, and $\tau_{en}$ are respectively:

$$\tau_{an} = 2\arctan(\frac{-E_n + \sqrt{H_n^2 + E_n^2 - D_n^2}}{H_n - D_n}) \quad (1)$$

$$\tau_{en} = 2\arctan(\frac{-J_n + \sqrt{K_n^2 + J_n^2 - I_n^2}}{K_n - I_n}) \quad (2)$$

where $D_n = -2X_{cn}l_2$, $E_n = -2Y_{cn}l_2$, and $H_n = l_2^2 + X_{cn}^2 + Y_{cn}^2 - l_3^2$. $I_n = -2(X_{cn} - l_1)l_5$, $J_n = -2Y_{cn}l_5$, and $K_n = l_5^2 + (X_{cn} - l_1)^2 + Y_{cn}^2 - l_4^2$.

In each of the three contact points, one force sensor was installed, as shown in Fig. 3. The independent control of the inverted five-bar linkages allows to generate different patterns at the user's palm and to move in different directions.

## III. TACTILE RENDERING

To provide high-fidelity tilt perception of the contact surfaces during the dexterous manipulation task, a high-efficiency digital signal processing is required. This section proposes a method to generate tactile patterns on the user's palm using the three contact points of the device. Two methods are proposed to convert the data from the tactile sensors on the gripper's fingers to the user's palm. The methods are described below.

### A. Downsize Tactile Patterns

The first stage of the method resizes and adapts the sensor data array (10x10 cells) to the dimension of the LinkGlide haptic display, reducing it to the stimulation array size (3x3 cells) with a unique stimulation point per row. We used the Bicubic Interpolation algorithm to downsize the tactile information and a maximum peak filter to assign the stimulation position for each contact point. This mathematical resampling algorithm produces a smooth output array with few interpolation artifacts [17] and without a relatively high computational cost. This stage results in digital tactile patterns that preserve most of their original contact surfaces' details and permit a more natural recognition of artificial tactile patterns.

### B. Mask Tactile Patterns using CNN Tilt Estimation

The second stage proposes to use a set of predefined tactile patterns as mask arrays (Fig. 5). The use of a specific masked pattern depends on the CNN tilt estimation. The final tactile stimuli will be the Boolean multiplication (AND) between the mask array and the downsized array generated in the previous stage. The final result of this operation will be the tactile stimulus arrays delivered by the multi-contact tactile display.

## IV. CNN-BASED PERCEPTION OF OBJECT ORIENTATION

### A. Architecture Description

To present a clear tactile pattern to the user, angle and position classification problems have to be solved. CNNs have found their applications in many areas, but commonly they are used for image processing. Nevertheless, it is possible to apply CNNs for tactile sensor data processing. In [18], failures during grasping were predicted by the information from an array of Inertial Measurement Units. Gandarias at al. [19] suggested to use CNN with high-resolution tactile sensors for object recognition.

For pipette angle and position recognition, we have implemented a classification CNN model with two heads, which architecture is shown in Fig. 6.

The proposed neural network consists of the backbone CNN network and two heads for angle and position classification. Backbone is responsible for feature extraction and consists of two convolutional layers with ReLU activation functions and batch normalization. Features are then fed into classification heads. Heads include four fully connected linear layers.

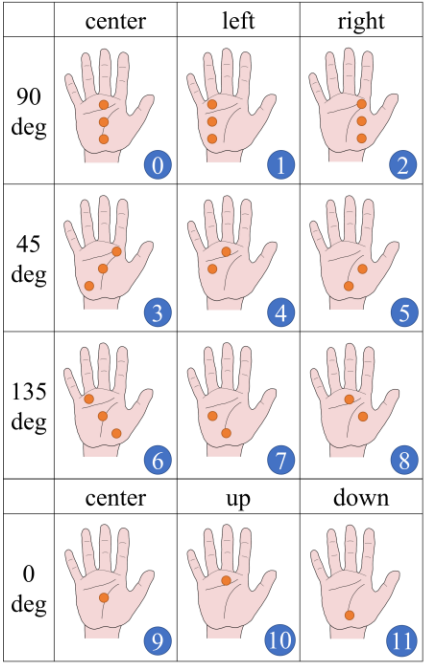

Fig. 5. Set of tactile pattern masks represented on the user's palm.

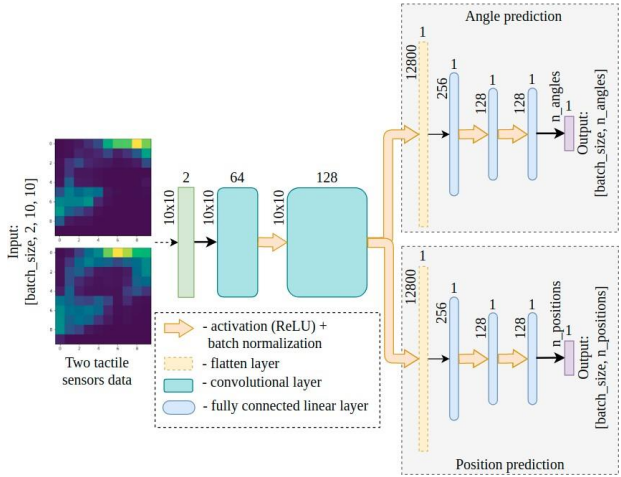

Fig. 6. CNN model architecture for angle and position classification from the tactile sensor data.

### B. Model Training and Validation

13392 data pairs from the tactile sensor arrays were collected for model training and validation. Angle classification problem included 4 classes (135 deg., 90 deg., 45 deg., and 0 deg.) and position classification problem included 3 classes (center, left and right pipette positions). 31 gripper positions were considered in the experiments (from minimum pressure applied to the pipette to maximum). The data was split into training set (50% of the dataset), validation set (25% of the dataset), and test set (25% of the dataset) for performance evaluation of the network. Data samples are shown in Fig. 7.

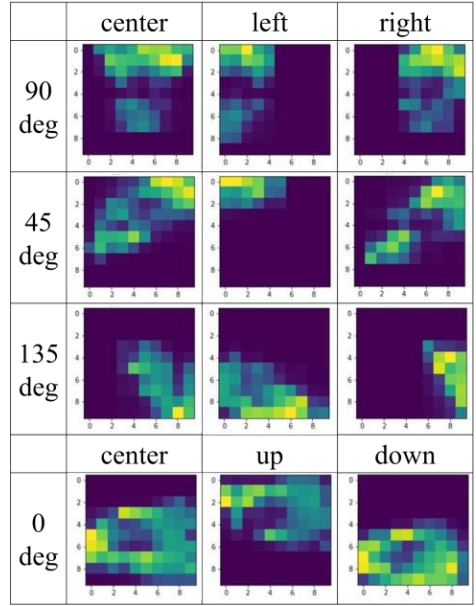

Fig. 7. Contact patterns for different positions and angles of pipette inclination.

Deep learning models were implemented with the PyTorch open-source machine learning framework. ReduceLROnPlateau scheduler was applied for dynamic learning rate reduction.

The cross-entropy loss was applied as a criterion for both classification heads. The total loss of the proposed neural network was the sum of the losses of two heads.

Angle prediction model achieved 95.09% test accuracy after 50 epochs of training. Position prediction model achieved 93.98 % test accuracy.

## V. EXPERIMENTAL EVALUATION

We conducted two user evaluations to determine the human perception of the orientation and position of the pipette during grasping by the two fingers gripper. The experiments evaluate the extent to which the tactile feedback with masked data, achieved by CNN classification, improves the users' perception at the palm of the deformable object tilt and orientation in comparison with the direct downsizing haptic feedback. The participants were informed about the experiments and agreed to the consent form. This study was approved by the Institutional Review Board of the Skolkovo Institute of Science and Technology.

## VI. EXPERIMENT ON TACTILE PERCEPTION

This evaluation is centered on the analysis of the human perception of the tactile rendering. During the first experiment, the tilt and position perception was rendered directly from sensors data to the LinkGlide display using the downsizing method described in Section III.A. During the second experiment, we assessed participants' perception of the pipette tilt angle and position with data masking performed by the CNN classification described in Section III.B.

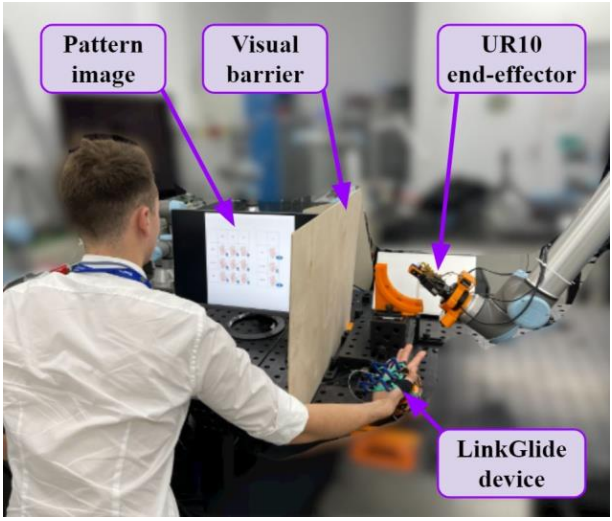

Fig. 8. The experimental setup of the multi-contact tactile perception during the grasping of the pipette by the 2-finger gripper.

Ten right-handed participants (4 females) aged 22 to 24 years volunteering complete the evaluation. None of them reported any deficiencies in sensorimotor function.

Before the study, the experimenter explained the purpose of the multi-contact haptic device to each participant and demonstrated the tactile feedback provided by the device for each of the four angles and three positions of the pipette. The demonstration was provided, at first, with additional visual feedback, and then at least one time blindly. During the experiment, the user was asked to sit in front of a desk and to wear the haptic display on the right hand as shown in Fig. 8. On the palm of the participants, the tactile patterns are rendered during the pipette grasping, evaluating the angle of its tilt and the position on the gripper. Each angle and position combination was presented 5 times blindly in random order, thus, 60 patterns were provided to each participant in each of the evaluation.

### A. Downsized Direct Data Rendering Results

The results of the human perception of the pipette tilt angle by rendering the downsized data are summarized in a confusion matrix (see Table I).

In order to evaluate the statistical significance of the differences between the perception of the angles and position without mask (12 patterns), we analyzed the results using single factor repeated-measures ANOVA, with a chosen significance level of $\alpha < 0.05$. The open-source statistical packages Pingouin and Stats models were used for the statistical analysis. According to the ANOVA results, there is a statistical significant difference in the recognition rates for the different angles, $F(11, 108) = 6.397$, $p = 4 \cdot 10^{-4}$. The ANOVA showed that the angles significantly influence the percentage of correct responses.

According to the ANOVA results, there is not statistical significant difference in the recognition rates for the different combinations of angles and positions, $F(11, 108) = 1.87$, $p = 0.051$. With these results, we can not confirm that statistically significant difference exists between the recognized patterns. The overall recognition rate is 9.67%, which means that the user can not distinguish the angles and positions. However, the overall recognition increases to 28% if only the angles are considered during the evaluation. The average recognition time is 4.97 sec.

### B. Masked Data Rendering Result

The results of the human perception of the pipette tilt angle and positions by the masked data rendering are summarized in a confusion matrix (see Table III).

TABLE I
CONFUSION MATRIX FOR ACTUAL AND PERCEIVED TILT ANGLES AND POSITIONS ACROSS ALL SUBJECTS FOR RECOGNITION WITHOUT MASKS.

| % | Answers (Predicted Class) | | | | | | | | | | | |
|---|---|---|---|---|---|---|---|---|---|---|---|---|
| Patterns | 0 | 1 | 2 | 3 | 4 | 5 | 6 | 7 | 8 | 9 | 10 | 11 |
| 0 | 0.16 | 0.02 | 0.18 | 0.06 | 0.04 | 0.12 | 0.06 | 0.04 | 0.04 | 0.10 | 0.08 | 0.10 |
| 1 | 0.06 | 0.06 | 0.32 | 0.12 | 0.00 | 0.08 | 0.00 | 0.06 | 0.00 | 0.10 | 0.10 | 0.10 |
| 2 | 0.14 | 0.02 | 0.16 | 0.14 | 0.02 | 0.14 | 0.04 | 0.00 | 0.10 | 0.14 | 0.06 | 0.04 |
| 3 | 0.12 | 0.02 | 0.22 | 0.14 | 0.04 | 0.12 | 0.02 | 0.00 | 0.04 | 0.06 | 0.16 | 0.06 |
| 4 | 0.12 | 0.04 | 0.16 | 0.16 | 0.00 | 0.20 | 0.00 | 0.02 | 0.08 | 0.08 | 0.04 | 0.10 |
| 5 | 0.08 | 0.06 | 0.18 | 0.12 | 0.08 | 0.10 | 0.06 | 0.06 | 0.04 | 0.14 | 0.06 | 0.02 |
| 6 | 0.18 | 0.04 | 0.20 | 0.06 | 0.02 | 0.08 | 0.06 | 0.08 | 0.06 | 0.08 | 0.12 | 0.02 |
| 7 | 0.16 | 0.02 | 0.16 | 0.12 | 0.08 | 0.10 | 0.02 | 0.02 | 0.12 | 0.16 | 0.02 | 0.02 |
| 8 | 0.12 | 0.02 | 0.22 | 0.08 | 0.08 | 0.04 | 0.04 | 0.08 | 0.12 | 0.02 | 0.06 | 0.12 |
| 9 | 0.08 | 0.02 | 0.24 | 0.10 | 0.00 | 0.12 | 0.10 | 0.04 | 0.14 | 0.08 | 0.04 | 0.04 |
| 10 | 0.10 | 0.10 | 0.32 | 0.06 | 0.08 | 0.04 | 0.00 | 0.04 | 0.10 | 0.06 | 0.08 | 0.02 |
| 11 | 0.10 | 0.02 | 0.06 | 0.04 | 0.08 | 0.08 | 0.10 | 0.04 | 0.12 | 0.02 | 0.16 | 0.18 |

TABLE II
CONFUSION MATRIX FOR ACTUAL AND PERCEIVED TILT ANGLES AND POSITIONS ACROSS ALL SUBJECTS FOR RECOGNITION USING MASKS.

| % | Answers (Predicted Class) | | | | | | | | | | | |
|---|---|---|---|---|---|---|---|---|---|---|---|---|
| Patterns | 0 | 1 | 2 | 3 | 4 | 5 | 6 | 7 | 8 | 9 | 10 | 11 |
| 0 | 0.92 | 0.00 | 0.04 | 0.04 | 0.00 | 0.00 | 0.00 | 0.00 | 0.00 | 0.00 | 0.00 | 0.00 |
| 1 | 0.02 | 0.78 | 0.00 | 0.00 | 0.12 | 0.02 | 0.00 | 0.04 | 0.02 | 0.00 | 0.00 | 0.00 |
| 2 | 0.04 | 0.00 | 0.80 | 0.00 | 0.00 | 0.14 | 0.02 | 0.00 | 0.00 | 0.00 | 0.00 | 0.00 |
| 3 | 0.04 | 0.02 | 0.02 | 0.76 | 0.06 | 0.04 | 0.00 | 0.04 | 0.02 | 0.00 | 0.00 | 0.00 |
| 4 | 0.12 | 0.02 | 0.02 | 0.00 | 0.18 | 0.58 | 0.06 | 0.00 | 0.00 | 0.04 | 0.00 | 0.00 |
| 5 | 0.00 | 0.00 | 0.02 | 0.00 | 0.00 | 0.96 | 0.00 | 0.00 | 0.02 | 0.00 | 0.00 | 0.00 |
| 6 | 0.08 | 0.00 | 0.00 | 0.00 | 0.00 | 0.00 | 0.92 | 0.00 | 0.00 | 0.00 | 0.00 | 0.00 |
| 7 | 0.06 | 0.02 | 0.00 | 0.00 | 0.00 | 0.00 | 0.02 | 0.90 | 0.00 | 0.00 | 0.00 | 0.00 |
| 8 | 0.00 | 0.00 | 0.10 | 0.00 | 0.02 | 0.08 | 0.08 | 0.02 | 0.66 | 0.02 | 0.00 | 0.02 |
| 9 | 0.00 | 0.00 | 0.00 | 0.06 | 0.00 | 0.00 | 0.02 | 0.00 | 0.86 | 0.00 | 0.06 | |
| 10 | 0.00 | 0.00 | 0.00 | 0.00 | 0.00 | 0.02 | 0.00 | 0.02 | 0.08 | 0.88 | 0.00 | |
| 11 | 0.02 | 0.00 | 0.00 | 0.00 | 0.00 | 0.00 | 0.00 | 0.00 | 0.00 | 0.10 | 0.00 | 0.88 |

In order to evaluate the statistical significance of the differences between the perception of the six angles, we analyzed the results using single factor repeated-measures ANOVA, with a chosen significance level of $\alpha < 0.05$. According to the ANOVA results, there is a statistical significant difference in the recognition rates for the different combination angles and positions, which correspond to the patterns in Fig. 5, $F(12, 108) = 2.190$, $p = 0.019$. The ANOVA showed that the angles and positions significantly influence the percentage of correct responses. The paired t-tests showed statistically significant differences between the pattern 0 and 4 ($p = 0.017 < 0.05$), 0 and 8 ($p = 0.025 < 0.05$), 10 and 4 ($p = 0.042 < 0.05$), 11 and 4

($p = 0.031 < 0.05$), 3 and 5 ($p = 0.034 < 0.05$), 4 and 5 ($p = 0.005 < 0.05$), 4 and 6 ($p = 0.011 < 0.05$), 4 and 7 ($p = 0.018 < 0.05$), 5 and 8 ($p = 0.021 < 0.05$), and 6 and 8 ($p = 0.044 < 0.05$). The open-source statistical package Pingouin was used for the statistical analysis.

The overall recognition rate is 82.5%. The pattern with less recognition rate is the number 4, which corresponds to 45 degrees on the left of the gripper. The pattern with higher recognition rate is the pattern number 5, which corresponds to 45 degrees to the right of the gripper. This effect can be caused because by the shape of the hand, having a better contact closer to the thumb. The average recognition time is 3.13 sec.

## VII. Conclusions and Future Work

In the presented work, the tactile tilt and position recognition of deformable objects was studied. The tilt and position data from the gripper while plastic pipettes were grasped was rendered to the users' palm by a multi-contact haptic device LinkGlide.

The experiments have shown that, the users' tilt and position perception of multi-contact tactile feedback with downsized data is significantly poor, with an average recognition rate of 9.67%, and only tilt perception of 28%. Applying the masked tactile patterns by CNN, we were able to increase the operator's perception of the tilt and position to 82.5%. Based on this evidence, we can conclude that the use of multi-modal tactile feedback on the users' palm in combination with the CNN-based rendering methods can potentially improve the telemanipulation of deformable objects.

Additionally, we can observe from the results, that the recognition rate for 45 deg. at the right is the higher perceived pattern (96%), followed by 0 and 135 deg. at the center (92%). However, the lower recognition rate was for 45 deg. at the left (58%). These results can be used to explore different methods to render the information combining different patterns.

The proposed system and CNN-based rendering method can be applied to increase the tilt and position recognition of laboratory instruments at remote co-working Labs, improving the dexterous telemanipulation and the users' response.

## Acknowledgment

The reported study was funded by RFBR and CNRS according to the research project No. 21-58-15006.